\documentclass[10pt,twocolumn,letterpaper]{article}

\usepackage{iccv}
\usepackage{times}
\usepackage{epsfig}
\usepackage{graphicx}
\usepackage{amsmath}
\usepackage{amssymb}

\usepackage{booktabs}

\usepackage{multirow}
\usepackage{algorithm}
\usepackage{algorithmic}
\usepackage[symbol]{footmisc}
\usepackage{threeparttable}
\usepackage{colortbl}
\usepackage{xcolor}

\usepackage[pagebackref,breaklinks,colorlinks]{hyperref}

\newcommand{\tablestyle}[2]{\setlength{\tabcolsep}{#1}\renewcommand{\arraystretch}{#2}\centering\footnotesize}
\newlength\savewidth
  
\usepackage[capitalize]{cleveref}
\crefname{section}{Sec.}{Secs.}
\Crefname{section}{Section}{Sections}
\Crefname{table}{Table}{Tables}
\crefname{table}{Tab.}{Tabs.}

\iccvfinalcopy 

\ificcvfinal\fi

\begin{document}


\title{FULLER: Unified Multi-modality Multi-task 3D Perception via Multi-level Gradient Calibration}

\makeatletter
\newcommand{\printfnsymbol}[1]{%
  \textsuperscript{\@fnsymbol{#1}}%
}
\makeatother

\author{Zhijian Huang$^{1}$\thanks{Equal contribution.} \quad Sihao Lin$^{2}$\printfnsymbol{1}  \quad Guiyu Liu$^{3}$\printfnsymbol{1}  \quad Mukun Luo$^{4}$ \quad Chaoqiang Ye$^{3}$ \quad Hang Xu$^{3}$\\ Xiaojun Chang$^{5,6}$ \quad Xiaodan Liang$^{1,6}$\thanks{Corresponding author.}\\
{\normalsize $^{1}$Shenzhen Campus of Sun Yat-sen University \quad $^{2}$RMIT University \quad $^{3}$Huawei Noah's Ark Lab}\\
{\normalsize $^{4}$Shanghai Jiao Tong University \quad $^{5}$University of Technology Sydney \quad $^{6}$MBZUAI}\\
{\tt\small huangzhj56@mail2.sysu.edu.cn, \{linsihao6,guiyuliou,chromexbjxh,cxj273,xdliang328\}@gmail.com, }\\
{\tt\small  luomukun@sjtu.edu.cn, yechaoqiang@huawei.com}
}

\maketitle
\begin{abstract}
Multi-modality fusion and multi-task learning are becoming trendy in 3D autonomous driving scenario, considering robust prediction and computation budget. However, naively extending the existing framework to the domain of multi-modality multi-task learning remains ineffective and even poisonous due to the notorious modality bias and task conflict.
Previous works manually coordinate the learning framework with empirical knowledge, which may lead to sub-optima.
To mitigate the issue, we propose a novel yet simple multi-level gradient calibration learning framework across tasks and modalities during optimization. 
Specifically, the gradients, produced by the task heads and used to update the shared backbone, will be calibrated at the backbone's last layer to alleviate the task conflict. Before the calibrated gradients are further propagated to the modality branches of the backbone, their magnitudes will be calibrated again to the same level, ensuring the downstream tasks pay balanced attention to different modalities.
Experiments on large-scale benchmark nuScenes demonstrate the effectiveness of the proposed method,  
\eg, an absolute 14.4\% mIoU improvement on map segmentation and 1.4\% mAP improvement on 3D detection,
advancing the application of 3D autonomous driving in the domain of multi-modality fusion and multi-task learning. We also discuss the links between modalities and tasks.
\end{abstract}

\vspace{-2mm}
\section{Introduction}
\label{sec:intro}

3D perception task plays an important role in autonomous driving. Previous works are mainly developed on single modality~\cite{yin2021center,li2022bevdepth, kendall2018multi, ye2022lidarmultinet, feng2021simple,philion2020lift,li2022bevformer, xie2022m,liu2022petr, liu2022petrv2} and different perception tasks are separated into individual models~\cite{ bai2022transfusion, li2022hdmapnet, lang2019pointpillars, yin2021multimodal, chen2022futr3d}. Often we wish to leverage complementary modalities to produce robust prediction and integrate multiple tasks within a model for the sake of computation budget. For instance, with the development of hardware, it is affordable to deploy both LiDAR and camera on a car, which are responsible to provide spatial information and semantic information. Integrating semantic-complementary vision tasks within a framework would greatly facilitate the deployment of real-world application~\cite{caesar2020nuscenes}. 

\begin{figure}[t]
  \centering
   \includegraphics[width=1.0\linewidth]{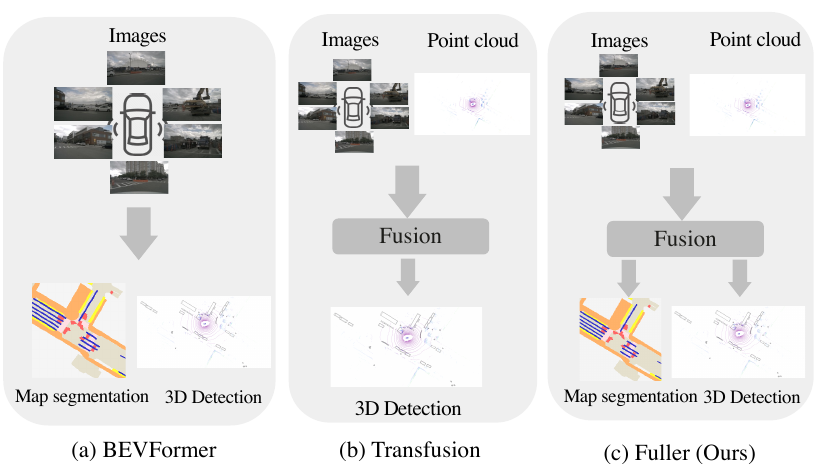}\vspace{-1mm}
   \caption{ \textbf{Comparison of paradigms on 3D perception.}
   (a) BEVFormer~\cite{li2022bevformer} focuses on multi-task learning, which could save the computation burden and thus facilitate the depolyment of real-world application.
   (b) Transfusion~\cite{bai2022transfusion} is proposed for multi-modality fusion for robust prediction since point cloud and images are complementary.
   (c) Our proposed Fuller is a unified framework that integrates these ingredients organically by solving the notorious problems of modality bias and task conflict.
   }\vspace{-5mm}
   \label{fig:motivation}
\end{figure}

Recent advances have stayed tuned for multi-modality fusion~\cite{liu2022bevfusion, liang2022bevfusion} and multi-task learning~\cite{zhang2022beverse, li2022bevformer} in the applications of 3D autonomous driving scenario. Meanwhile, it is of great interest to unify multi-modality fusion and multi-task learning within a framework. In fact, it is unlikely to expect that dumping all the individual components into one framework and they would function organically. We build up a competitive baseline based on BEVFusion~\cite{liu2022bevfusion}, which takes as input both the point cloud and image, and serves for two complementary vision tasks: 3D detection (foreground) and map segmentation (background). However, we observe the severe issues of modality bias and task conflict: a) different tasks prefer specific modality, \eg, 3D detection relies on spatial information provided by LiDAR sensor while segmentation task relies more on image inputs.
b) adding a new task will degrade both tasks: -3.0 \% mAP for detection and -18.3\% mIoU for map segmentation.

From the perspective of optimization, we investigate the potential gradient imbalance that occurs during end-to-end training in a hierarchical view. First, we study the gradients which are produced by different task heads and are applied to update the parameters of the shared backbone. We observe that simply summing up these raw gradients to update the shared backbone would damage the performance of both tasks, suggesting an imbalance between them. Empirical findings prove that there is a great discrepancy between the gradient magnitudes w.r.t. the task objectives. Second, we inspect the gradients produced in the intra-gradient layer, which is to be separated into successive modality branches. Given a trained baseline, we visualize the gradient distributions of different modality branches and find their magnitudes imbalanced greatly. We further calculate the task accuracy by dropping one of the modalities to measure the modality bias. Our findings catch up with the theoretical analysis of~\cite{wang20222modality}, which suggests that the point cloud and image branches are suffering from the imbalanced convergence rate w.r.t. the downstream tasks. 

We motivate our method by noting the findings discussed above, which is proposed to uni\textbf{f}y m\textbf{u}lti-modality mu\textbf{l}ti-task 3D perception via mu\textbf{l}ti-level gradi\textbf{e}nt calib\textbf{r}ation, dubbed as \textit{\textbf{Fuller}}. Specifically, we devise the multi-level gradient calibration, comprised of inter-gradient and intra-gradient calibration, to address the associated issues.
In terms of the task conflict, we find that the task with lower gradient magnitude would be overwhelmed by another task with higher gradient magnitude. Thus, we propose to calibrate the gradients of different task losses at the backbone. Since the gradient would be manipulated at the layer level, this technique is referred to as \textbf{inter-gradient} calibration.
Regarding modality bias, we expect the different modalities can update and converge at the same pace. Hence, before the gradients are separated into the modality branches, we calibrate their magnitudes to the same level, which is performed in the intra-gradient layer internally and thus called \textbf{intra-gradient} calibration.

On top of the gradient calibration, we introduce two lightweight heads for our tasks. These two heads are both transformer-based. With our specially designed initialization methods, they can generate fine-grained results with just a one-layer decoder, allowing to save much more parameters than dense heads.

We thoroughly evaluate the Fuller on the popular benchmark nuScenes~\cite{caesar2020nuscenes}.
Regarding the sensory input, we adopt the point cloud to provide accurate spatial information and use the image to compensate for the lack of visual semantics. In terms of perception tasks, we select two representative and complementary tasks: 3D detection and map segmentation, which are responsible for dynamic 
foreground objects and static road elements understanding.
Note that BEVFusion~\cite{liu2022bevfusion} only organizes these ingredients empirically without mentioning the problems discussed above. To summarize, our contributions are:


\begin{itemize}
    \vspace{-2mm}
    \item We propose the Fuller which organically integrates multi-modality fusion and multi-task learning for 3D perception via multi-level gradient calibration during end-to-end optimization.
    \vspace{-2mm}
    \item We introduce the new architecture design for task heads, which outperforms or is comparable with the previous head design while saving $\sim$40\% parameters.
    \vspace{-2mm}
    \item Extensive experiments demonstrate that Fuller can prevent the notorious problems of modality bias and task conflict, \eg, an absolute 14.4\% mIoU improvement on map segmentation and 1.4\% mAP improvement on 3D detection.
\end{itemize}

\begin{figure*}[t]
  \centering
    \includegraphics[width=1.0\linewidth]{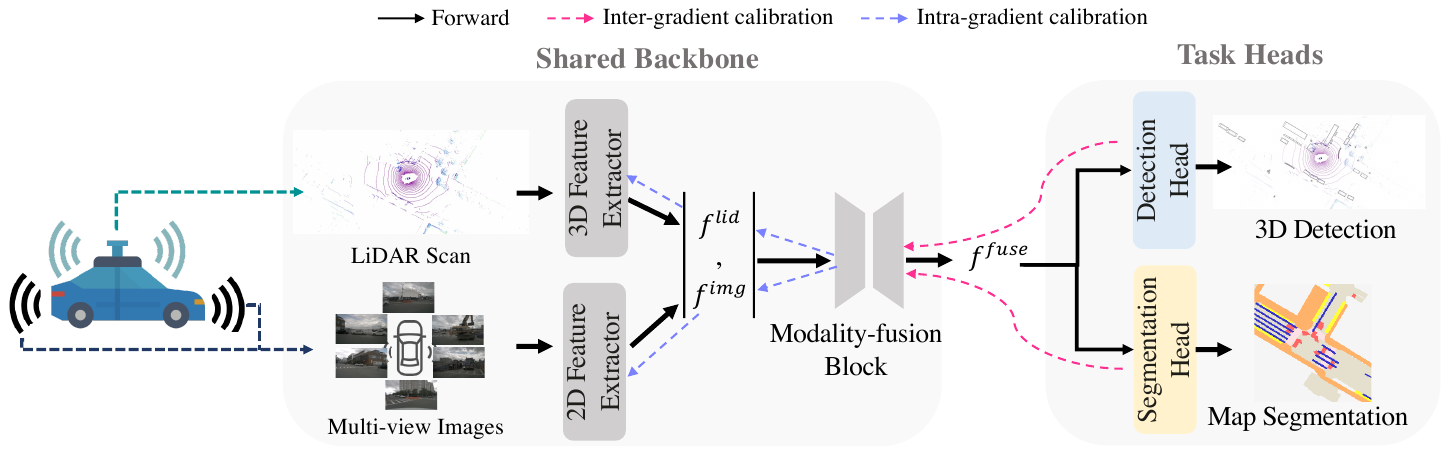}\vspace{-0mm}
  \caption{\textbf{Framework of the Fuller.}
  Generally, Fuller takes as input the LiDAR scan and multi-view images and predicts two tasks: 3D detection and map segmentation. 
  We propose multi-level gradient calibration to deal with the problems of task conflict and modality bias during optimization:
  i) The gradients, produced by the task heads and applied on the shared backbone, will be calibrated on the last layer of the backbone, namely, inter-gradient calibration (pink dashed line). 
  ii) When it comes to the subsequent modality branches of the shared backbone, the gradient magnitudes will be calibrated again to the same level within the intra-gradient layer, referred to as intra-gradient calibration (blue dashed line). We also introduce a lightweight design for the task heads, which saves $\sim$40\% parameters.}
  \label{fig:framework}\vspace{-5mm}
\end{figure*}
\vspace{-5mm}
\section{Related Work}
\label{sec:formatting}

\noindent \textbf{3D perception tasks in autonomous driving.}
Lidar and image are the two most powerful and widely used modalities in the area of autonomous driving.  Multimodal fusion has been well-studied to boost the performance of 3D object detection task\cite{vora2020pointpainting, yin2021multimodal,chen2022futr3d, bai2022transfusion, liang2022bevfusion}.
Multi-task networks of 3D perception also arouse significant interest in autonomous driving community. These multi-task studies are limited on uni-modal network architectures, either with a Lidar backbone\cite{kendall2018multi, ye2022lidarmultinet, feng2021simple} or an image backbone\cite{philion2020lift,li2022bevformer, xie2022m,liu2022petr, liu2022petrv2}. MMF\cite{liang2019multi} works on depth completion and object detection with both camera and LiDAR inputs, but depth estimation only works as an auxiliary head and only object detection was evaluated.
BEVFusion\cite{liu2022bevfusion} is the first multimodal network to perform object detection and map segmentation simultaneously. However, BEVFusion\cite{liu2022bevfusion} focuses on single task and network acceleration, and only provides two pieces of joint training results. Our proposed method is the first multimodal multitask network, and we evaluate each task and analyze them from the perspectives of multimodal and multitask.

\noindent \textbf{Multimodal learning.}
Multimodal learning is increasingly used to improve the performance of certain tasks, such as action recognition\cite{gao2020listen,ismail2020improving, kazakos2019epic}, visual question answering \cite{antol2015vqa, ilievski2017multimodal} and perception tasks in autonomous driving\cite{liu2022bevfusion, bai2022transfusion, liang2022bevfusion}.
Most multi-modality research focuses on the network structure, such as concatenation, convolution or gated fusion in the middle or later part of the network\cite{kiela2018efficient, owens2018audio, hu2018squeeze}. Few studies\cite{wang2020makes, peng2022balanced} concentrate on multimodal optimization methods during the training process.
\cite{wang2020makes} proposes a metric OGR to quantize the significance of overfitting and try to solve it with Gradient Blending. It designs modal heads for each task thus difficult to expand to multi-task network.
OGM-GE\cite{peng2022balanced} try to solve the optimization imbalance problem by dynamically adjusting the gradients of different modalities. 
Since it separates parameters of different modalities in the linear classification head, it is hard to generalize to other complicated task heads.
Differently, our method can be used in networks with any task head as long as the network has modal-specific parameters.

\noindent \textbf{Multi-task optimization methods.}
Multi-task methods are mainly divided into two categories in \cite{vandenhende2021multi}, network architecture improvement\cite{misra2016cross,xu2018pad, ruder2019latent,guo2020learning} and optimization methods\cite{liu2019end,chen2018gradnorm,liu2021towards,yu2020gradient,mordan2021detecting}. Our approach focuses on the optimization methods. The goal of multi-task optimization methods is to balance the loss weights of different tasks to prevent one task from overwhelming another during training. DWA\cite{liu2019end} adjusts the loss weights based on the rate at which the task-specific losses change, but it requires to balance the loss magnitudes beforehand.
Gradnorm\cite{chen2018gradnorm} balances the loss weights automatically by stimulating the task-specific gradients to be of similar magnitude. IMTL\cite{liu2021towards} optimizes the training process by guaranteeing the aggregated gradient has equal projections onto individual tasks. 
Yet they have not been studied in the domain of multi-modality multi-task learning. Our method complements these analysis.

\vspace{-2mm}
\section{Method}
\vspace{-1mm}
In this section, we introduce the Fuller, a framework that unifies the multi-modality multi-task 3D perception in autonomous driving scenarios. Fuller aims to mitigate the problem of modality bias and task conflict during the end-to-end training by gradient calibration.
Regarding the network architecture, we introduce a lightweight design for the task heads, named Fuller-det and Fuller-seg.

\subsection{Network architecture}
\label{sec:arch}
As shown in \cref{fig:framework}, our proposed Fuller extracts features from both point cloud and images, then transforms them into a unified bird's-eye view (BEV) representation. It relies on VoxelNet~\cite{zhou2018voxelnet} as LiDAR backbone and Swin-T~\cite{liu2021swin} as image backbone. 
As for image features from multi-view cameras, we project them onto BEV feature using the scheme as same as LSS~\cite{philion2020lift}.
We adopt the modality fusion strategy where the features of two branches, $f^{img}$ and $f^{lid}$, are first concatenated and then fed into the fusion block:
\begin{equation}
    f^{fuse} = {\texttt{conv}}(f^{lid}\oplus f^{img}),
\label{eq:fuse}
\end{equation}
\noindent where $\texttt{conv}$ is the modal fusion block (\ie, 2-layer FPN) and $\oplus$ is concatenation operation.
$f^{fuse}$ is then connected to task-specific heads. 

The detection head Fuller-det follows a DETR-style~\cite{carion2020end} architecture with object queries. 
Given the fusion featrure $f^{fuse}$, Fuller-det initializes the queries by an auxiliary heatmap head according to TransFusion\cite{bai2022transfusion}.
Also, Fuller-seg utilizes a query-based semantic segmentation head with segmentation queries. 
The BEV feature $f^{fuse}$ is transformed into the output shape feature $F$. 
The initialized queries and $F$ are then used to obtain mask embeddings $M$, processed by the transformer decoder layer. 
Finally, the binary mask prediction $S$ is computed via a dot product between $M$ and $F$, followed by a sigmoid activation.

We refer the reader to App. {\color{red}{B}} for more details.

Both Fuller-det and Fuller-seg have only one transformer decoder layer and could achieve competitive results compared to state-of-the-art methods, as will see in~\cref{sec:comparison}. 


\begin{figure}
  \centering
  \includegraphics[width=1\linewidth]{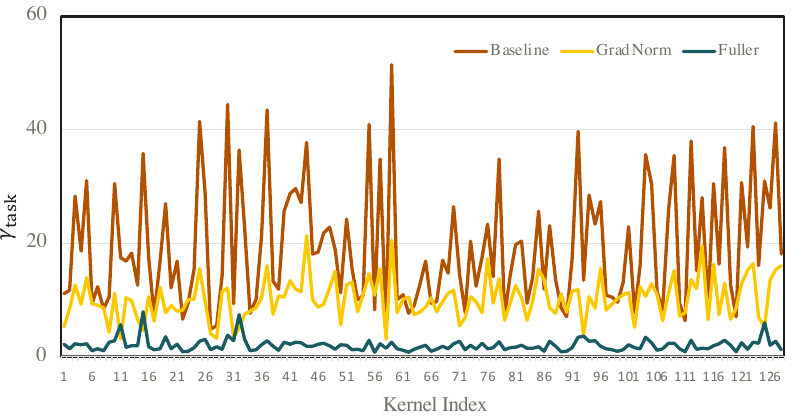}\vspace{-0mm}
  \caption{ We visulize the $\gamma_{\rm task}$ (\cref{eq:magnitude}) in the last layer of the modality-fusion block. The gradient tensors are unfolded along the first axis.
  It is easy to observe that the gradient magnitude of seg loss dramatically lags behind that of the det loss. We apply the proposed Fuller and compare it with GradNorm~\cite{chen2018gradnorm}. We find that our method is able to balance the gradients from two tasks. Importantly, our method yields a more stable and balanced $\gamma_{\rm task}$.
  }\vspace{-4mm}
  \label{fig:grad_magnitude_task}
\end{figure}

\subsection{Multi-level Gradient Calibration}
We now introduce the multi-level gradient calibration. First, it will calibrate the gradient between tasks via inter-gradient calibration. When it comes to the subsequent modality branches of the backbone, the gradient will be calibrated again by intra-gradient calibration.
\vspace{-3mm}
\subsubsection{Inter-Gradient Calibration for Task Conflict}
By definition, the gradients will be propagated from the task heads to the shared backbone.
Without any regularization, multi-task learning would simply sum up the individual gradients for backbone update. Since the gradients of the downstream tasks tend to exhibit great distinction, this naive manner will inevitably result in task conflict. For example, an objective with low gradient magnitude would be overwhelmed by another one with high gradient magnitude. Therefore, existing works~\cite{vandenhende2021multi,chen2018gradnorm,liu2019end,liu2021towards} propose to \textit{manipulate} the gradients to interfere the optimization process. 

Following this philosophy, we visualize the gradient distribution of the two tasks to inspect the inferior performance. 
Specifically, we compute the ratio of $L2$ norm between the gradients computed by raw individual losses:
\begin{equation}
    \gamma_{\rm task} = \frac{||\nabla_{\texttt{shared\_L}}\mathcal{L}_{\rm{Det}}||}{||\nabla_{\texttt{shared\_L}}\mathcal{L}_{\rm{Seg}}||},
\label{eq:magnitude}
\end{equation}
\noindent where $\nabla$ denotes gradient computation operator, $\mathcal{L}_{\rm Det}$ and $\mathcal{L}_{\rm Seg}$ are the output losses of 3D detection and map segmentation, respectively. 
Typically, the gradients of shared backbone computed by different task losses are utilized to measure task characteristics. To save computation, we select the last layer of shared backbone, denoted as $\texttt{shared\_L}$. 
Thus, $\gamma_{\rm task}$ is a metric that reflects the gradient discrepancy.

As we might notice in~\cref{fig:grad_magnitude_task}, the value of $\gamma_{\rm task}$ between the two tasks is significantly huge. Based on this finding, we consider that the emergence of task conflict is probably because the gradients of segmentation task are overwhelmed by that of detection task. Inspired by the loss weighting methods~\cite{vandenhende2021multi,chen2018gradnorm,liu2019end,liu2021towards}, we balance the gradients of different tasks by balancing their loss weights. At each iteration, we obtain the gradients corresponding to individual loss on the last layer of the shared backbone.
These gradients are utilized to derive the new loss weights. Then, the aggregated loss is applied to calibrate the gradients of the entire network.
We evaluate existing literature and choose the IMTL\_G~\cite{liu2021towards} as the technique for this purpose given its superior performance, 
as discussed in App. {\color{red}{C.2}}.
\vspace{-3mm}
\subsubsection{Intra-Gradient Calibration for Modality Bias}
We have analyzed the impact of different task objectives on the backbone holistically. Another complicated situation arises when optimizing modality branches. During our experiments, we observe the issue of modality bias which undermines the assumption that multiple modalities can collaboratively support the downstream tasks. This phenomenon is also known as semantic inconsistency~\cite{goel2022cyclip} and modality imbalance~\cite{wang20222modality}.

The first layer of the modality fusion block is referred to as the intra-gradient layer, parameterized by $\theta^{F}$. It consists of two parts, $\theta^{F}_{lid}$ and $\theta^{F}_{img}$, that represent the parameters directly connected to the LiDAR and image backbones during backpropagation. Let $H$ denotes the modality branches, where $\theta^H_{lid}$ and $\theta^H_{img}$ represent the parameters of the LiDAR and image branch. 
According to the chain rule, 
the gradient for a certain modality branch is defined as:

\begin{equation}
G_{mod} = \frac{\partial\mathcal{L}}{\partial{\theta_{mod}^{H}}}     
= \frac{\partial \mathcal{L}}{\partial{\theta^{F}_{mod}}}                \cdot
            \frac{\partial {\theta^{F}_{mod}}}{\partial{\theta_{mod}^{H}}}, 
\label{eq:grad_sensor}
\end{equation}

\noindent where $mod$ = $\{lid, img\}$, $G_{lid}$ and $G_{img}$ mean the gradients of the two modality branches.
According to~\cref{eq:grad_sensor}, $\nabla{\theta^{F}_{lid}}=\frac{\partial \mathcal{L}}{\partial \theta^{F}_{lid}}$ 
would carry out the updating message from the task heads to the LiDAR branch, similarly for
image branch.

Regarding the term 
$\nabla_{\theta^{F}}\mathcal{L}$, 
this gradient corresponds to the optimization process that determines how the intra-gradient layer would coordinate the fusion of the two modalities to adapt to downstream tasks. Therefore, we use $\nabla{\theta^{F}_{lid}}$ and $\nabla{\theta^{F}_{img}}$ within the intra-gradient layer to establish the connection between two modality branches. 
Since
they will be separated into different branches, we consider their relative magnitude during end-to-end training:


\begin{equation}
        \gamma_{\rm modal} = \frac{||\nabla{\theta^{F}_{lid}}||}{||\nabla{\theta^{F}_{img}}||}.
\label{eq:grad_magnitude_modal}
\end{equation}

\begin{figure}[!t]
  \centering
  \includegraphics[width=1.0\linewidth]{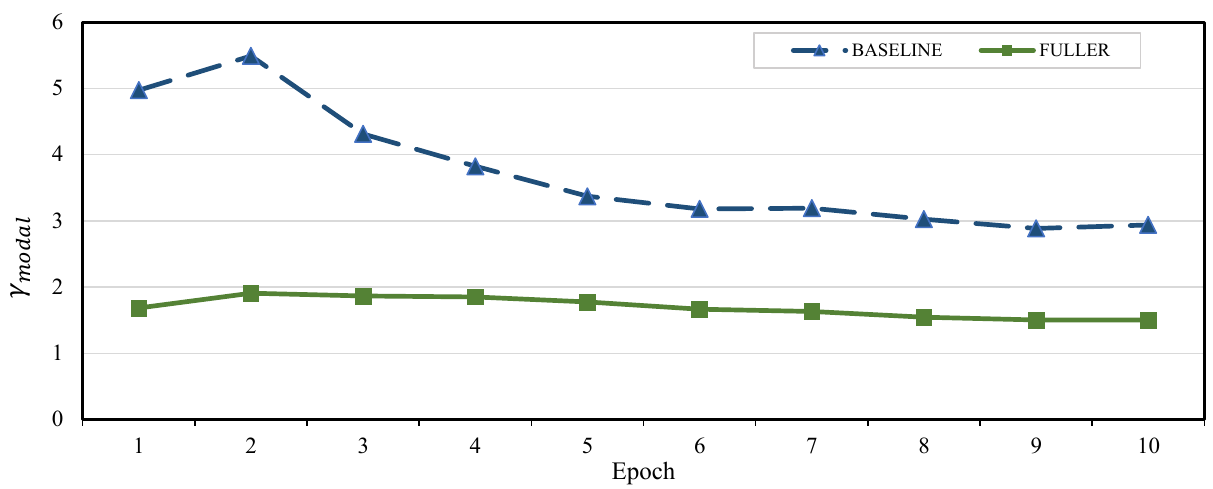}\vspace{-0mm}
  \caption{Compared with baseline, Fuller has a balanced $\gamma_{\rm modal}$ (\cref{eq:grad_magnitude_modal}), meaning that two modalities can be learned in a balanced manner.}
  \label{fig:grad_magnitude_modal}\vspace{-5mm}
\end{figure}

The result displayed in~\cref{fig:grad_magnitude_modal} indicates that for most of the time, $||\nabla{\theta^{F}_{lid}}||$ would surpass $||\nabla{\theta^{F}_{img}}||$, which means the LiDAR and image branches receive uneven attention from the downstream tasks. 


To solve this problem, we propose calibrating the gradients between two branches, \ie, $G_{lid}$ and $G_{img}$. In practice, we gate the one with greater magnitude to slow down its pace, ensuring that the tasks pay balanced attention to both modalities. At $t$ step, we obtain the gating factors by:


\begin{equation}
\begin{aligned}
   w^{t}_{lid} &= \sigma(||\nabla{{\theta^{F}_{lid}}^{t}}||,||\nabla{{\theta^{F}_{img}}^{t}}||)\in(0,1],\\
   w^{t}_{img} &= \sigma(||\nabla{{\theta^{F}_{img}}^{t}}||,||\nabla{{\theta^{F}_{lid}}^{t}}||)\in(0,1],
\label{eq:gating_factor}
\end{aligned}
\end{equation}

\begin{gather}
    \sigma(x,y)=\textbf{1}_{\frac{x}{y}>1}(1- \texttt{tanh}(\alpha\cdot \frac{x}{y}))+\textbf{1}_{\frac{x}{y}<=1},
\label{eq:alpha}
\end{gather}
\noindent where $\sigma(\cdot,\cdot)$ is a composition function conditioned by the indicator function and used to measure a paired input. $\alpha$ is a weight factor.
The gating factors in~\cref{eq:gating_factor} are further smoothed by momentum update with coefficient $m$ to stabilize the training. Then the calibrated gradient will be backpropagated to the associated branch:
\begin{equation}
\begin{aligned}
        w^{t}_{mod} &= m\cdot w^{t-1}_{mod} + (1-m)\cdot w^{t}_{mod},
\end{aligned}
\label{eq:momentum}
\end{equation}
\begin{equation}
    G^{t}_{mod} = w^{t}_{mod} \cdot G^{t}_{mod}.
\label{eq:cali_grad}
\end{equation}

\noindent We refer this technique as intra-gradient calibration where it is performed 
between modalities.

\subsection{Fuller: The Blueprint}

We have presented a hierarchical view of the inter-gradient and intra-gradient calibration techniques, which are proposed to optimize the entire backbone and the associated modality branches, respectively. The procedure of Fuller is summarized in~\cref{alg:fuller}. At each update step, we first calculate the gradients w.r.t. the two objectives in the last layer of the shared backbone.
The two gradients are calibrated to alleviate the problem of task conflict, where a pair of weights are derived. 
After applying the weights to the raw losses, we obtain the calibrated gradient of the total loss on the intra-gradient layer, 
$\nabla{\theta^{F}_{lid}}$ and $\nabla{\theta^{F}_{img}}$. To mitigate the issue of modality bias, 
we utilize them to calibrate the gradients of corresponding branches.

\begin{algorithm}[t]
    \renewcommand{\algorithmicrequire}{\textbf{Input}}
	\renewcommand{\algorithmicensure}{\textbf{Output}}
    \caption{Training Procedure of Fuller}
    \label{alg:fuller}
    \begin{algorithmic}[1]
        \REQUIRE 
        composition function $\sigma$, 
        modality branches' parameter $\theta^H$, 
        intra-gradient layer's parameter $\theta^F$,
        off-the-modality's parameter $\theta^K$, learning rate $\eta$.
        \ENSURE $\theta^H$, $\theta^K$
        \FOR { $t=0,1,2,...,T$}
        \STATE Inter-gradient Calibration For Task Conflict: \\ 
        
        
        {\small $\alpha_{\rm Seg}$, $\alpha_{\rm Det}$
        $\leftarrow$ {\rm IMTL}(
        $\nabla_{\texttt{shared\_L}}\mathcal{L}_{\rm{Det}},
        \nabla_{\texttt{shared\_L}}\mathcal{L}_{\rm{Seg}}$)}

        \STATE $\mathcal{L}\leftarrow \alpha_{\rm Seg}\cdot \mathcal{L}_{\rm Seg} + \alpha_{\rm Det}\cdot \mathcal{L}_{\rm Det}$
        \STATE $\theta^K \leftarrow \theta^K - \eta \cdot \frac{\partial \mathcal{L}}{\partial \theta^K}$ \ $\triangleleft$ \ Update task heads
        \STATE Intra-gradient Calibration for Modality Bias: \\
        
        \STATE $\nabla{\theta^{F}_{lid}},  \nabla{\theta^{F}_{img}} \leftarrow \frac{\partial \mathcal{L}}{\partial \theta^{F}_{lid}}, \frac{\partial \mathcal{L}}{\partial \theta^{F}_{img}}$
        
        
        \STATE $w^{t}_{lid} \leftarrow \sigma(||\nabla{{\theta^{F}_{lid}}^{t}}||,||\nabla{{\theta^{F}_{img}}^{t}}||)$\\
        $w^{t}_{img} \leftarrow \sigma(||\nabla{{\theta^{F}_{img}}^{t}}||,||\nabla{{\theta^{F}_{lid}}^{t}}||)$
        \STATE $w^{t}_{lid} \leftarrow m\cdot w^{t-1}_{lid} + (1-m)\cdot w^{t}_{lid}$ \\ 
        $w^{t}_{img} \leftarrow m\cdot w^{t-1}_{img} + (1-m)\cdot w^{t}_{img}$ 
        \STATE $G^{t}_{lid} \leftarrow w^{t}_{lid} \cdot G^{t}_{lid}$
        ; \; $G^{t}_{img} \leftarrow w^{t}_{img} \cdot G^{t}_{img}$
        \STATE backward $G^{t}_{lid}$ and update $\theta^{H}_{lid}$ \; \; $\triangleleft$ LiDAR branch
        \STATE backward $G^{t}_{img}$ and update $\theta^{H}_{img}$ $\triangleleft$ Image branch
        \ENDFOR
    \end{algorithmic}
\end{algorithm}

\section{Experiments}
\vspace{-0mm}
\label{sec:exp}
We first introduce our baseline setting and benchmark dataset. 
We also investigate the potential of Fuller by evaluating it under different loss weights and dataset distribution settings.
Finally, we ablate the proposed components to validate their individual effectiveness.
\vspace{-1mm}
\subsection{Experimental Settings}
\vspace{-1mm}
\noindent \textbf{Implementation details.}
We adopt BEVFusion~\cite{liu2022bevfusion} as the strong baseline with a few modifications. The detailed design has been discussed in~\cref{sec:arch}. 
The AdamW optimizer is utilized with a weight decay of $10^{-2}$ and momentum of 0.9.
The models are trained for 10 epochs with a learning rate of $10^{-3}$.
We use 8 NVIDIA V100 GPUs with 2 samples per GPU, resulting in a total batch size of 16. 
Additionally, the value of $\alpha$ used in~\cref{eq:alpha} is set to 0.1, while $m$ in~\cref{eq:momentum} is set to 0.2.

\noindent \textbf{nuScenes dataset.} nuScenes~\cite{caesar2020nuscenes} is a multi-sensor dataset that provides diverse annotations for multiple tasks, including detection, tracking, and especially BEV map segmentation, which is typically absent in other datasets. 
The dataset comprises 28,130 training samples and 6,019 validation samples, each containing a 32-beam LiDAR scan and 6 multi-view images. 
The 3D detection task involves 10 foreground categories, and the performance is evaluated by mean Average Precision (mAP) and nuScenes Detection Score (NDS). For map segmentation, the model is required to segment 6 background categories in BEV view, which is measured by the mean Intersection over Union (mIoU). 

\noindent \textbf{Evaluation protocol.}
\label{sec:dataset}We evaluate the performance of multi-task learning 
based on the metric in~\cite{vandenhende2021multi}:

\begin{equation}
\Delta_{\rm MTL} = \frac{(-1)^{l}}{T}\sum_{i=1}^{T}(M_{m,i}-M_{b,i})/M_{b,i},
\label{eq:mtl}
\end{equation}
\noindent
where $T$ is the number of tasks.
$M_{m,i}$ and $M_{b,i}$ are the performance of the $i$-th task of the evaluated model and baseline, respectively.
$\Delta_{\rm MTL}$ could be intuitively understood as the average performance drop,
where we set $l =1$, \ie, lower value means better performance.
Following Liang \etal~\cite{liang2022effective}, we also evaluate the Fuller in three annotation schemes.

\noindent \textbf{Full setting.} 
We leverage all available annotations by default, which serves as the upper bound for the following two settings.

\noindent \textbf{Disjoint-normal.} 
Given the limited budget, the annotation complexity determines the quantity of task labels. In a realistic practice, we split the full dataset into 3D detection and map segmentation parts using a 3:1 ratio, where each sample is labeled for one task.


\noindent \textbf{Disjoint-balance.} 
Similarly, each sample is endowed with a task label and each task can leverage half of the dataset.

\begin{table}[t]
    \centering
    \caption{Sensitivity analysis and ablation study of the proposed gradient calibration with different initial loss weights.}
    \vspace{-0mm}
    \resizebox{1.0\columnwidth}{!}
    {\tablestyle{8pt}{1.0}
    \begin{tabular}{@{}cc|cccc@{}}
    \toprule
        Intra. & Inter. &mAP(\%)$\uparrow$ &NDS$\uparrow$ &mIoU(\%)$\uparrow$ 
        &{ $\Delta_{\rm MTL}$}(\%)$\downarrow$\\
        \midrule
        \multicolumn{6}{c}{\textit{det\_weight:seg\_weight=1:1}}\\
        \midrule
        &  &59.1 & 65.0 & 44.0 &18.3\\
        \checkmark & &59.5 &\textbf{65.4} &45.0 &16.9\\
        &\checkmark & 57.1 &63.3 &\textbf{59.5} &8.8\\
        \checkmark &\checkmark &\textbf{60.5}  &65.3  &58.4 &\textbf{5.4}\\
        
        \midrule
        \multicolumn{6}{c}{\textit{det\_weight:seg\_weight=1:5}}\\
        \midrule
        &  &59.8 & 65.5 & 55.7 &8.0\\
        \checkmark &  &59.8 &65.3 &56.1 &7.8 \\
        &\checkmark &56.9 &63.3  &\textbf{59.8} &8.7 \\
        \checkmark &\checkmark & \textbf{60.1} & \textbf{65.6} & 58.2 &\textbf{5.7}\\
        
        \midrule
        \multicolumn{6}{c}{\textit{det\_weight:seg\_weight=1:10}}\\
        \midrule
        &  &59.3 & 65.0 & 57.9 &7.0\\
        \checkmark & & \textbf{60.1} & \textbf{65.4} & 57.3 &6.5\\
        &\checkmark &58.2  &64.2 & \textbf{60.1} &6.7\\
        \checkmark &\checkmark & 59.9 &65.2 &59.2 &\textbf{5.3}\\
\bottomrule
    \end{tabular}}
    \label{tab:ablation}\vspace{-4mm}
\end{table}
\subsection{Sensitivity Analysis}
\noindent \textbf{Initial states.} 
To investigate the robustness of our method w.r.t. initial loss weights, we incrementally increase the weight of segmentation loss and inspect its impact on performance. 
As illustrated in~\cref{tab:ablation}, the loss weights between detection and segmentation are set to 1:1, 1:5, and 1:10. 
We find that increasing the loss weight of map segmentation can improve the subsequent performance of the baseline model. However, manually adjusting the loss weights can lead to sub-optimal results. For instance, when the loss weight of map segmentation is increased from 5 to 10, it benefits map segmentation (55.7\%$\rightarrow$57.9\% mIoU) but damages the performance of detection task (59.8\%$\rightarrow$59.3\% mAP). Nonetheless, the proposed method can facilitate model training despite variations in initial loss weights.

\begin{table}[!t]
    \centering
    \caption{Sensitivity analysis and ablation study of the proposed gradient calibration under different dataset distribution.}
    \resizebox{1.0\columnwidth}{!}
    {\tablestyle{8pt}{1.0}
    \begin{tabular}{@{}cc|cccc@{}}
    \toprule
        Intra. & Inter. &mAP(\%)$\uparrow$ &NDS$\uparrow$ &mIoU(\%)$\uparrow$ 
        &{ $\Delta_{\rm MTL}$}(\%)$\downarrow$\\
        \midrule
        \multicolumn{6}{c}{\textit{Full}}\\
        \midrule
         & &59.1 & 65.0 & 44.0 &18.3\\
        \checkmark & &59.5 &\textbf{65.4} &45.0 &16.9\\
        &\checkmark & 57.1 &63.3 &\textbf{59.5} &8.8\\
        \checkmark &\checkmark &\textbf{60.5}  &65.3  &58.4 &\textbf{5.4}\\
        \midrule
        \multicolumn{6}{c}{\textit{Disjoint-balance}}\\
        \midrule
         & &\textbf{58.7} & 64.2 & 41.5 &21.3\\
        \checkmark &  &58.3 &\textbf{65.0} &42.4 &20.5 \\
        &\checkmark &57.4 &62.7  &\textbf{57.3} &10.8 \\
        \checkmark &\checkmark & 58.4 & 63.6 & 56.7 &\textbf{9.8}\\
        \midrule
        \multicolumn{6}{c}{\textit{Disjoint-normal}}\\
        \midrule
        &&59.1 & 64.7 & 43.1 &19.3\\
        \checkmark & &\textbf{59.3} &\textbf{65.3}  &44.0  &17.9\\
        &\checkmark &57.4  &62.9 & 53.8 &13.4\\
        \checkmark &\checkmark & 58.9 &64.9 &\textbf{55.0} &\textbf{9.7}\\
\bottomrule
    \end{tabular}}
    \label{tab:disjoint_ablation}
    \vspace{-4mm}
\end{table}
\noindent \textbf{Dataset distributions.} 
We verify the model under different dataset distributions~\cite{liang2022effective}.
In~\cref{tab:disjoint_ablation}, the baseline performance of disjoint dataset is notably inferior compared with that of full dataset,
posing a greater challenge to multi-task learning. 
In the full setting, Fuller was observed to improve both tasks. Given the significant $\Delta_{\rm MTL}$ metric (42.5\%) of the disjoint-balance baseline, Fuller would pay more attention to improving map segmentation while slightly degrading 3D detection to address task conflict. In the case of disjoint-normal, the baseline's $\Delta_{\rm MTL}$ (38.5\%) is relatively small and has minor effect on 3D detection. Generally, Fuller can boost $\Delta_{\rm MTL}$ metric under these scenarios.

\begin{table*}[t]
    \centering
    \caption{Comparison with benchmark. The upper two sub-tables are single task results while the bottom one is multi-task result.‘L’ and ‘C’ represent LiDAR and Camera, respectively. We treat single task result as our upper bound because multi-task will generally decrease the performance. Baseline means Fuller is naively trained where detection loss and segmentation loss are set to 1:1. `-' means inapplicable.
    }\vspace{-0mm}
    \resizebox{2.0\columnwidth}{!}
    {\tablestyle{8pt}{1.0}
    \begin{threeparttable}

    \begin{tabular}{@{}l|cccc|ccc}
    \toprule
          &Modality& VoxelSize & LiDAR &Image &mAP(\%)$\uparrow$ &NDS$\uparrow$ &mIoU(\%)$\uparrow$ \\
          \midrule
          \midrule
         \multicolumn{8}{c}{3D Detection}\\
          \midrule
         BEVFormer~\cite{li2022bevformer} & C & -& -& ResNet101~\cite{He_2016_CVPR}&\cellcolor{blue!5}41.6 &\cellcolor{blue!5}51.7 & -\\
         CenterPoint~\cite{yin2021center} &L & 0.075&VoxelNet &- &\cellcolor{blue!5}59.6 &\cellcolor{blue!5}66.8 &- \\
         MVP$^{\ddag}$~\cite{yin2021multimodal}& C+L & 0.075 &  VoxelNet & DLA-34& \cellcolor{blue!5}66.1 & \cellcolor{blue!5}70.0 & -\\
         TransFusion~\cite{bai2022transfusion} & C+L &0.075 & VoxelNet&DLA-34& \cellcolor{blue!5}67.5 & \cellcolor{blue!5}71.3 & - \\
         BEVFusion~\cite{liu2022bevfusion} & C+L &0.075 & VoxelNet&Swin-T &  \cellcolor{blue!5}68.5 & \cellcolor{blue!5}71.4 & - \\
         Fuller-det &C+L  &0.075 & VoxelNet&Swin-T & \cellcolor{blue!5}\textbf{67.6} &\cellcolor{blue!5}\textbf{71.3}  &- \\
         Fuller-det (upper bound) &C+L  &0.1 &VoxelNet&Swin-T & \cellcolor{blue!5}\textbf{62.1} &\cellcolor{blue!5}\textbf{66.6} &- \\
         \midrule
          \midrule
         \multicolumn{8}{c}{BEV Map Segmentation}\\
          \midrule
         LSS$^{\ddag}$~\cite{philion2020lift} &C  &- &- &EfficientNet-B0&- &-&\cellcolor{blue!5}44.4 \\
         CenterPoint$^{\ddag}$~\cite{yin2021center} &L &0.1  &VoxelNet &- &- &-  &\cellcolor{blue!5}48.6 \\
         BEVFusion~\cite{liu2022bevfusion} & C+L  &0.1 & VoxelNet&Swin-T & -&- & \cellcolor{blue!5}62.7 \\
         Fuller-seg(upper bound) &C+L  &0.1 &VoxelNet&Swin-T&- & - &\cellcolor{blue!5}\textbf{62.3} \\
         \midrule
          \midrule
         \multicolumn{8}{c}{3D Detection + BEV Map Segmentation}\\
          \midrule
         BEVFusion$^{\dag}$~\cite{liu2022bevfusion} (share) & C+L  &0.1 & VoxelNet&Swin-T &\cellcolor{blue!5}- & \cellcolor{blue!5}69.7 & \cellcolor{blue!5}54.0\\
         BEVFusion$^{\dag}$~\cite{liu2022bevfusion} (sep) & C+L  &0.1 &VoxelNet&Swin-T & \cellcolor{blue!5}- & \cellcolor{blue!5}69.9 & \cellcolor{blue!5}58.4\\
         Baseline(share)& C+L  &0.1 &VoxelNet&Swin-T& \cellcolor{blue!5}59.1 & \cellcolor{blue!5}65.0 & \cellcolor{blue!5}44.0\\
         Fuller(share) &C+L  &0.1 & VoxelNet&Swin-T &\cellcolor{blue!5}\textbf{60.5} &\cellcolor{blue!5}\textbf{65.3}  &\cellcolor{blue!5}\textbf{58.4}\\
         \bottomrule
    \end{tabular}
    \begin{tablenotes}
        \footnotesize
        \item[$\dag$] means the multi-task result in BEVFusion\cite{liu2022bevfusion}. $^{\ddag}$ means re-implementation result in BEVFusion\cite{liu2022bevfusion}. `share' means multi-task heads share one BEV encoder to process the fused multimodal feature. `sep' means task heads have separate encoders.
    \end{tablenotes}
\end{threeparttable}}
\vspace{-3mm}
\label{tab:baseline}
\end{table*}

\begin{table}[t]
  \centering
  \vspace{-1mm}
  \caption{
  Relation between inter- and intra-gradient calibration.
  }\vspace{-0mm}
\resizebox{0.9\columnwidth}{!}
{\tablestyle{12pt}{1.0}
  \begin{tabular}{@{}cc|ccc@{}}
    \toprule
        Intra. & Inter. &$\gamma_{\rm task}$$\downarrow$ 
        &$\gamma_{\rm modal}$$\downarrow$ &$\Delta_{\rm MTL}$ $\downarrow$\\
        \midrule
        & &19.2 &3.7 &18.3 \\
        \checkmark & &12.5 &1.9 &16.9 \\
         &\checkmark &2.5  &2.8 &8.8 \\
        \checkmark &\checkmark &\textbf{1.9}  &\textbf{1.7} &\textbf{5.4} \\
\bottomrule
    \end{tabular}}
  \label{tab:relation_inter_intra}
  \vspace{-4mm}
\end{table}

\subsection{Ablation Study}
\noindent \textbf{Validating the inter- and intra-gradient calibration.} We conduct thorough experiments, including three different initial states, to validate the individual effectiveness of the proposed multi-level gradient calibration in~\cref{tab:ablation}. 
Generally, the intra-gradient calibration leads to considerable improvements in the downstream tasks compared to the baseline. 
Regarding the inter-gradient calibration, it can largely enhance the performance of the map segmentation while deteriorating the detection task at the acceptable cost.
The combination of the two techniques yields remarkable improvement in both tasks, ultimately achieving best $\Delta_{\rm MTL}$. 
We further evaluate the validity of the two proposed components across various dataset distribution, as shown in~\cref{tab:disjoint_ablation}. Again, both components can individually improve the $\Delta_{\rm MTL}$ in all settings.

\noindent \textbf{Relation between two calibration techniques.} 
The above experiments (\cref{tab:ablation}\&\cref{tab:disjoint_ablation}) evidence the individual effectiveness of the two proposed calibration techniques. We are curious whether the calibrations are consistently cooperative 
or could be adversarial in certain scenarios.
To investigate this, we visualize the $\gamma_{\rm modal}$ and $\gamma_{\rm task}$ by ablating one of the calibration techniques, as shown in~\cref{tab:relation_inter_intra}. Interestingly, we found that applying either calibration technique could simultaneously mitigate both issues of modality bias and task conflict, resulting in more balanced $\gamma_{\rm modal}$ and $\gamma_{\rm task}$. The result indicates that the two proposed calibration techniques are cooperative.

\subsection{More Results}
\label{sec:comparison}
\begin{figure}[!t]
  \vspace{-1mm}
  \centering
    \includegraphics[width=1.0\linewidth]{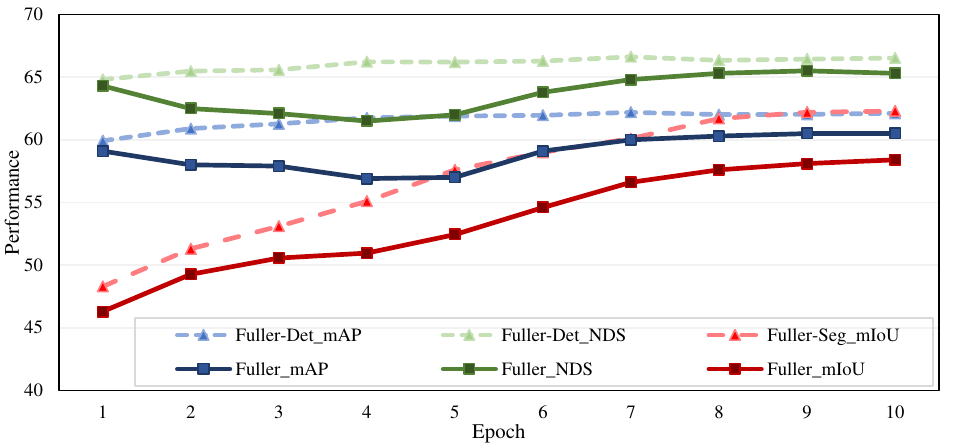}
  \caption{Model convergence of single-task and multi-task models.}
  \label{fig:model_convergence}\vspace{-6mm}
\end{figure}

\vspace{-1mm}
\noindent \textbf{Comparison with the benchmark.}
We compare the Fuller with current state-of-the-art methods and report the result on nuScenes validation set (\cref{tab:baseline}). We list each model's modality and group them by task setting. 
Our baseline (\ie, penultimate row) is adapted from the competitive BEVFusion~\cite{liu2022bevfusion}.
Given hardware capacity of V100 GPU, the voxel size is set to 0.1m for multi-task learning.
For fair comparison, the single-task model Fuller-det and Fuller-seg using voxel size of 0.1m are set as the upper bounds for 3D detection and map segmentation. 
As shown in~\cref{fig:model_convergence}, Fuller-det converges after 7 epochs with lr=1e-4. Fuller-seg converges after 10 epochs with lr=1e-3.  We train the Fuller using the scheme as same as Fuller-seg.

As illustrated in~\cref{tab:baseline}, the performance of multi-task baseline suffers a significant decline compared to the upper bounds.
Particularly, the mIoU of segmentation task drops drastically from 62.3\% to 44.0\%, which discourages the multi-task applications in autonomous driving scenarios. 
Rather, the proposed model Fuller demonstrates substantial improvement in bridging the gap between single-task and multi-task models, 
which improves mIoU from 44.0\% to 58.4\% in map segmentation and facilitates the mAP from 59.1\% to 60.5\% in 3D detection.

\begin{figure*}[t]
  \centering
  \includegraphics[width=1.0\linewidth]{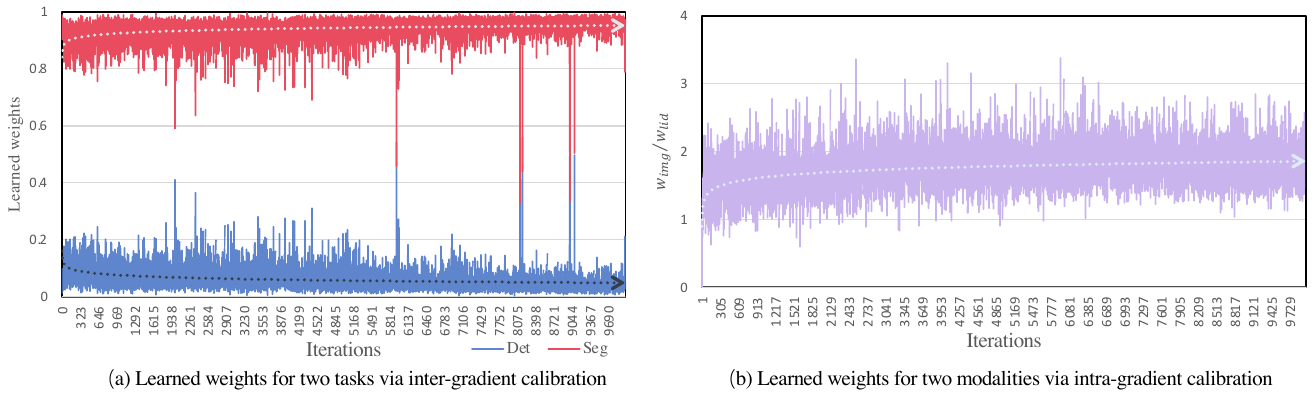}\vspace{-0mm}
  \caption{Visualization of gradient calibration.}
  \label{fig:vis_fuller}\vspace{-4mm}
\end{figure*}


\noindent \textbf{Visualization of multi-level gradient calibration.}
\label{sec:vis_fuller}
To better understand the mechanism of the proposed multi-level gradient calibration, we visualize how Fuller manipulates the gradients during network optimization. 
As shown in~\cref{fig:vis_fuller} (a), Fuller adapts its loss functions to mitigate task conflict by decreasing the detection loss and elevating the segmentation loss.
In terms of modality bias, we plot the ratio $w_{img}/w_{lid}$ (\cref{eq:gating_factor}) for visualization. According to \cref{fig:vis_fuller} (b), the weight $w_{lid}$ is smaller than $w_{img}$ for most of the time, indicating that Fuller would gate the gradients of LiDAR branch to take maximum advantage of both modalities, thereby improving the subsequent result. 

\begin{table}[t]
    \centering
  \caption{Memory cost and inference speed. All the speeds are evaluated on an Tesla V100 GPU.}
\resizebox{1.0\columnwidth}{!}
{
\tablestyle{22pt}{1.0}
  \begin{tabular}{@{}cccc@{}}
    \toprule
    & Memory$\downarrow$ & Parameter$\downarrow$ & FPS$\uparrow$ 
     \\
     \midrule
    STL & 6144MB & 81.97M & 1.20  \\
    FULLER & 3103MB &  44.12M & 2.30 \\
    \bottomrule
  \end{tabular}}\vspace{-3mm}
  \label{tab:efficiency}
\end{table}
\begin{table}[t]
  \centering
  \vspace{-0mm}
  \caption{Verifying the framework with more learning tasks.}
  \vspace{-0mm}
\resizebox{1.0\columnwidth}{!}
{\tablestyle{1pt}{1.0}
  \begin{tabular}{@{}lccccc@{}}
    \toprule
    Method  &mAP(\%)$\uparrow$ & NDS$\uparrow $&mIoU$_{\rm map}$(\%)$\uparrow$ & mIoU$_{\rm fore}$(\%)$\uparrow$ &$\Delta_{\rm MTL}$ $\downarrow$\\
    \midrule
    Upper bound &62.1 &66.6 &62.3 &63.8 &-\\
    MTL baseline &\textbf{59.9} &\textbf{65.8} & 46.9 & 58.1 &12.8\\
    Fuller &58.6 & 64.6 &\textbf{57.1} &\textbf{62.0} &\textbf{9.9}\\
    \bottomrule
  \end{tabular}}\vspace{-5mm}
  \label{tab:add_new_task}
\end{table}

\begin{figure}[t]
  \centering
  \includegraphics[width=1.0\linewidth]{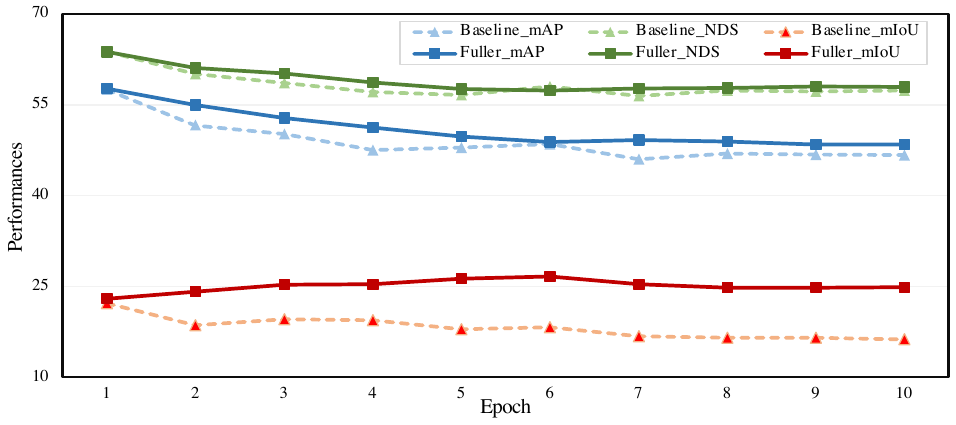}\vspace{-1mm}
  \caption{A model trained with both modalities is evaluated by dropping off the image input. LiDAR can still support the detection task. However, without image input, the performance on map segmentation drops largely.
  Fuller proves to be able to mitigate the modality bias and  significantly improves map segmentation result.}
  \label{fig:drop_img}
  \vspace{-6mm}
\end{figure}

\noindent \textbf{Association between task and modality.}
To identify the association between modalities and tasks, we propose evaluating the \textit{trained} model with one modality removed at a time. Specifically, we examine the performance of the Fuller and baseline models in the absence of image input (\cref{fig:drop_img}). The results indicate that 3D detection retains a considerable level of accuracy even without image input, thanks to the precise spatial information provided by LiDAR scans. In contrast, the absence of image input significantly impairs the performance of map segmentation.
Our findings are consistent with the theoretical analysis in~\cite{peng2022balanced}, which suggests that each modality carries out a unique mechanism and contributes distinct functionality to downstream tasks. 
In App. {\color{red}{C.5}}, we provide additional experiments and discussion in which LiDAR scans are absent.

\noindent \textbf{Generalization ability.}
By sharing a common backbone, Fuller can save substantial memory cost and speed up the inference as shown in~\cref{tab:efficiency}.
Additionally, we augment our framework to 3 tasks by introducing foreground segmentation, as demonstrated in~\cref{tab:add_new_task}. 
Foreground segmentation is a task related to 3D detection and map segmentation, which is segmentation of foreground objects under BEV.
Our proposed approach still achieves performance gains.


\vspace{-3mm}
\section{Conclusion}
\vspace{-2mm}
We present Fuller, a framework that addresses the challenges of modality bias and task conflict in multi-modality multi-task learning for 3D perception tasks. To cope with these problems, we propose multi-level gradient calibration to guide the learning process of the model. Our approach includes inter-gradient calibration to balance the gradients w.r.t. downstream tasks on the last layer of the shared backbone. Before being separated into different branches, the magnitude of these gradients will be calibrated again within the intra-gradient layer.

\vspace{-3mm}
\section{Acknowledgements}
\vspace{-3mm}
This work was supported in part by National Key R\&D Program of China under Grant No. 2020AAA0109700,  Guangdong Outstanding Youth Fund (Grant No. 2021B1515020061), Shenzhen Science and Technology Program (Grant No. RCYX20200714114642083)
Shenzhen Fundamental Research Program(Grant No. JCYJ20190807154211365), Nansha Key RD Program under Grant No.2022ZD014 and Sun Yat-sen University under Grant No. 22lgqb38 and 76160-12220011. We thank MindSpore for the partial support of this work, which is a new deep learning computing framwork\footnote{https://www.mindspore.cn/}.

   
\clearpage

{\small
\bibliographystyle{ieee_fullname}
\bibliography{iccv2023.bbl}
\
}

\end{document}